\documentclass[letterpaper, 10 pt, conference]{ieeeconf}  

\IEEEoverridecommandlockouts                              

\usepackage{graphicx}
\usepackage{amsmath,amssymb,amsfonts}
\usepackage{hyperref}
\usepackage{gensymb}
\usepackage{tabularx}
\usepackage{booktabs}
\usepackage{lipsum}
\usepackage{multirow}
\usepackage{threeparttable}
\hypersetup{hidelinks}

\title{\LARGE \bf
Towards Over-Canopy Autonomous Navigation: Crop-Agnostic LiDAR-Based Crop-Row Detection in Arable Fields}

\author{ Ruiji Liu$^{1}$, Francisco Yandun$^{1}$ and George Kantor $^{1}$ 
\thanks{$^{1}$Robotics Institutes, Carnegie Mellon University, US
{\tt\small {\{ruijil, fyandun, gkantor\}@andrew.cmu.edu}}}
}

\begin{document}
\maketitle
\thispagestyle{empty}
\pagestyle{empty}

\begin{abstract}
Autonomous navigation is crucial for various robotics applications in agriculture. However, many existing methods depend on RTK-GPS devices, which can be susceptible to loss of radio signal or intermittent reception of corrections from the internet. Consequently, research has increasingly focused on using RGB cameras for crop-row detection, though challenges persist when dealing with grown plants. This paper introduces a LiDAR-based navigation system that can achieve crop-agnostic over-canopy autonomous navigation in row-crop fields, even when the canopy fully blocks the inter-row spacing. Our algorithm can detect crop rows across diverse scenarios, encompassing various crop types, growth stages, the presence of weeds, curved rows, and discontinuities. Without utilizing a global localization method (i.e., based on GPS), our navigation system can perform autonomous navigation in these challenging scenarios, detect the end of the crop rows, and navigate to the next crop row autonomously, providing a crop-agnostic approach to navigate an entire field. The proposed navigation system has undergone tests in various simulated and real agricultural fields, achieving an average cross-track error of $3.55cm$ without human intervention. The system has been deployed on a customized UGV robot, which can be reconfigured depending on the field conditions.
\end{abstract}

\section{Introduction}
The escalating global population, labor shortage, and environmental challenges pose significant concerns regarding food shortage. These challenges emphasize the urgent need for sustainable agricultural practices and innovative solutions to ensure food security for future generations \cite{food_shortage}. In this scenario, agricultural robots have emerged as an attractive solution to tackle these challenges in agri-food production, offering precision farming capabilities \cite{Human-robot}. 

Across agricultural robotics, robust and safe autonomous navigation without damaging the crops plays a pivotal role in different applications. Existing technologies can safely drive a robot and perform various tasks on the pre-devised path by utilizing the Real Time Kinematic Global Positioning System (RTK-GPS) \cite{RTK-GPS1}\cite{RTK-GPS2}\cite{RTK-GPS3}\cite{RTK-GPS4}. However, the deployment of RTK-GPS devices on robots can be cost-prohibitive for many farmers, particularly when accurate crop row positions are unavailable due to traditional planting methods. In addition, the RTK network coverage and the obstacles above the GPS antenna will affect the accuracy of the receiving position corrections. In the worst-case scenario, loss of GPS signal, even for a brief moment, can lead to a malfunction of the navigation system, causing the robot to damage the crops and pose a potential danger in general. These issues motivate the development of precise crop-row detection algorithms that leverage local sensors such as cameras and LiDAR. By having precise crop row measurements, the robot can follow the crop rows, achieving resilient autonomous navigation independent of the robot's global position. Most previous research focuses on camera-based crop-row detection due to the abundant visual information cameras provide. Nevertheless, field conditions in reality, such as crop types, crop growth stages, discontinuities, weeds presence, illumination, and curved rows, pose significant challenges to precise crop-row detection and robot localization with existing crop-row detection algorithms. 
\begin{figure}[t]
    \centering
    \includegraphics[width = 0.48\textwidth]{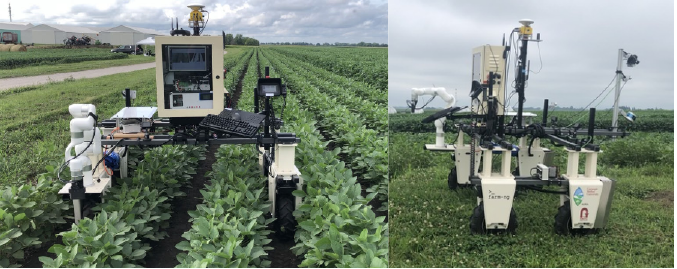}
    \caption{Amiga robot navigating within soybean fields (left) and robot configuration (right). Our navigation system is a LiDAR-based autonomous navigation system for over-canopy navigation in row-crop fields.}
    \vspace{-1em}
    \label{fig:introduction1}
\end{figure}

In this paper, we introduce a LiDAR-based over-canopy crop-row detection algorithm combined with a crop-row following and lane-switching function for autonomous navigation. Our system enables robots to navigate on different crops and cover the entire field without human intervention. The main contributions of our work are summarized below:
\begin{itemize}
  \item A novel and robust multi-crop-row detection algorithm designed to operate effectively in fields with varying crop types, growth stages, the presence of weeds, curved rows, and discontinuities.
  \item A multi-crop-row following and lane-switching systems integrated with a nonlinear local Model Predictive Control (MPC) algorithm for autonomous full field coverage.
  \item Field validation of the complete autonomous navigation system on a full-scale robot. We conducted tests on simulated corn and soybean fields, as well as real corn fields at various growth stages (young\footnote{Crop height: $0.2\text{m}$-$0.4\text{m}$, open canopies} and grown\footnote{Crop height: $0.4\text{m}$-$0.7\text{m}$, semi-closed or closed canopies}).
\end{itemize}

\begin{figure*}
\begin{center}
\includegraphics[width=0.99\textwidth]{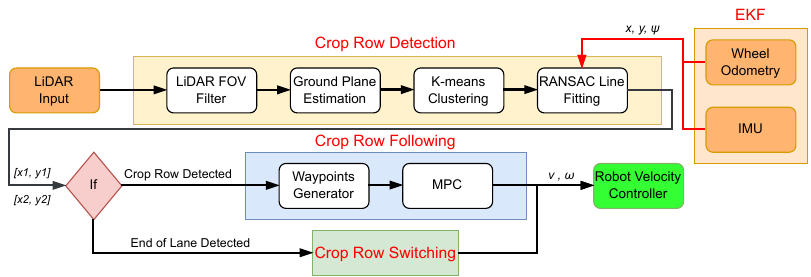}
\end{center}
\caption{Navigation system's workflow. (\romannumeral 1) The crop row detection algorithm uses LiDAR data and filtered odometry values $[x, y, \psi]$ as inputs, predicting crop rows in the form $[x1, y1, x2, y2]$ within the robot's frame. (\romannumeral 2) The crop row following algorithm applies nonlinear MPC to control the robot to follow the center line of the predicted rows, sending linear velocity $v$ and angular velocity $w$ commands. (\romannumeral 3) The crop row switching algorithm utilizes a PID controller to navigate the robot to the next lane if no more crop rows are detected.}
\vspace{-1em}
\label{fig:system}
\end{figure*}

\section{Related Work}
Autonomous agricultural robots play a crucial role in modern farming, performing tasks such as seeding \cite{seeding}, harvesting \cite{harvesting}, and crop monitoring \cite{monitoring} to improve productivity and sustainability. The development of fully autonomous navigation systems for agricultural robots, capable of operating across diverse crop fields, is an active area of research. Although GPS with RTK corrections is widely utilized for autonomous navigation in outdoor farm environments \cite{RTK-GPS1}\cite{RTK-GPS2}\cite{RTK-GPS3}\cite{RTK-GPS4}, its application is usually constrained by getting accurate crop positions and limitations in signal coverage. Consequently, significant research efforts have been put into crop-row detection algorithms as a more reliable alternative for autonomous navigation within agricultural fields. 
\begin{figure}[b]
\vspace{-1em}
    \centering
    \includegraphics[width =0.48\textwidth]{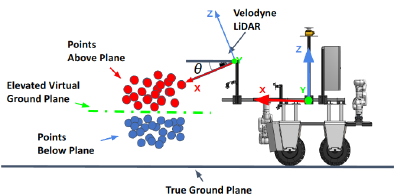}
    \caption{LiDAR point clouds filtering techniques. The virtual ground plane is estimated based on the LiDAR tilted angle $\theta$ and intersected with the centroids of the filtered LiDAR data. Points below the plane (blue) are removed, resulting in simplified LiDAR data (red) for crop-row detection.}
    \label{fig:plane}
\end{figure}

Previous research on crop-row detection has employed various methods and combinations of onboard sensors. RGB cameras are frequently used for crop-row detection due to the rich visual information they provide. English \textit{et al.} \cite{texture}, Guerrero \textit{et al.} \cite{texture1}, and Ronchetti \textit{et al.} \cite{texture2} explored texture extraction from RGB images, while Ahmadi \textit{et al.} \cite{ahmadi2020visual}\cite{multi_crop_row} utilized color information to create crop row masks and predict row locations. In addition, the advent of deep learning has introduced new approaches for crop-row detection, with recent studies employing image segmentation techniques \cite{deep-learning1}\cite{deep-learning2}\cite{deep-learning3}\cite{deep-learning4}. However, most of these camera-based crop-row detection algorithms are only tested on a limited type of crop fields and early stages of development, while the deep learning methods even require new training when navigating in unseen scenarios. Furthermore, these methods struggle when dense canopy covers the inter-row spacing.

Additionally, LiDAR has also been utilized for crop-row detection as an alternative sensor to RGB cameras. Malavazi \textit{et al.} \cite{Lidar-based} developed a method for extracting line parameters from 2D LiDAR point clouds but only in orchard-like configurations, while Baquero \textit{et al.} \cite{Baquero_Velasquez_2022} utilized 3D LiDAR point for under-canopy navigation. Winterhalter \textit{et al.}'s work on crop-row detection \cite{lidar_on_tiny} demonstrated the ability to detect various crop sizes and types, but it lacked validation for real-time autonomous navigation in field conditions. More sensor fusion methods \cite{multi-sensor}\cite{multi-sensor1} have been proposed to address challenging field scenarios. However, similarly to camera-based methods, none of these works offer robust solutions for over-canopy navigation on crops with fully blocked canopies, such as grown corn or soybean. 

To address the issues mentioned above, we propose an innovative crop-agnostic crop-row detection algorithm utilizing a 3D LiDAR. Our crop-row detection method has undergone several tests in both real and simulated corn and soybean fields (including different growth stages, discontinuities, weeds, and curved rows). Additionally, we integrate the crop-row following algorithm with a lane-switching system, allowing robots to navigate an entire field autonomously. The complete navigation system was deployed and tested on both real and simulated Amiga robot. We have open-sourced this navigation system and simulation environments \footnote{\url{https://github.com/Kantor-Lab/LiDAR_CropRowDetection}}.

\section{System Design}
Fig. \ref{fig:system} illustrates our navigation system workflow. Our navigation system consists of three parts: crop-row detection, crop-row following, and crop-row switching. The crop-row detection algorithm takes LiDAR data and filtered odometry values $[x, y, \psi]$, estimated from an EKF fusing wheel odometry and IMU \cite{robot-localization}, and publishes the prediction of crop rows as lines $[x1, y1, x2, y2]$ in the local frame. The row-following algorithm then takes the crop row positions and generates waypoints along the center line. It follows the crop rows by controlling the robot's linear velocity $v$ and angular velocity $w$ using a nonlinear MPC approach. If the end of the crop lane is detected, the robot will perform a lane-switching maneuver to enter the next lane.

\subsection{Robotic Platform}
\label{sec:robot_platform}
For all of our tests, we used a customized version of the Amiga robot, manufactured by Farm-Ng \cite{farm-ng}. Fig. \ref{fig:introduction1} provides an overview of this vehicle and the working environments. This robot is an all-electric robot platform that can be configured to perform various tasks in different crops. With a 250 lbs base weight, this platform provides a 1000 lbs payload for 8 hours on flat ground, combined with a maximum speed of 5.5 mph. For navigation purposes, the width of the platform was set to $1.8m$ to accommodate the $0.75m$ inter-row spacing with a vertical clearance of $0.88m$. The length of the platform was set to $1.2m$ to provide enough space for mounting sensors and other hardware. For evaluating the navigation system, we installed a SwiftNav GPS device with RTK correction for ground truth, one Vectornav IMU, and one Velodyne VLP 16 LiDAR in the center front of the robot. During the navigation, we obtain the robot's real-time position and orientation in 2D, $[x, y, \psi]$, using an EKF  \cite{robot-localization} with only IMU and wheel odometry inputs. All software runs onboard on a customized PC with a mini-ITX motherboard, an Intel i9 13900 CPU, and an NVIDIA GeForce RTX 4070 within a weather-resistant box \cite{guri2024hefty}. Future work envisions a robotics arm interacting with the vegetation based on the sensor feedback from above the canopy. Therefore, an over-canopy strategy was chosen for navigation and manipulation.

\subsection{Overall Navigation Strategy}\label{sec:strategy}
The ultimate goal of our navigation system is to autonomously drive the robot through the entire crop field while performing various tasks (e.g., monitoring or manipulation). The robot starts within the crop rows and autonomously follows them utilizing the LiDAR-based navigation system. Upon detecting the end of the lane, the robot starts the lane-switching process. For this, we implemented a PID controller method that uses mainly filtered odometry data. Since the crops are typically planted in parallel rows for efficient management, this approach first rotates the robot by 90 degrees, drives it forward based on the distance between crop rows, and then performs another 90-degree turn to navigate into the next lane. Then, the LiDAR-based navigation system is started again and navigates the robot through the new lane. This process is implemented as a finite state machine, which enables the automatic change of states during execution.

\begin{figure}[t]
\vspace{0.2cm}
  \centering
  \includegraphics[width=0.48\textwidth]{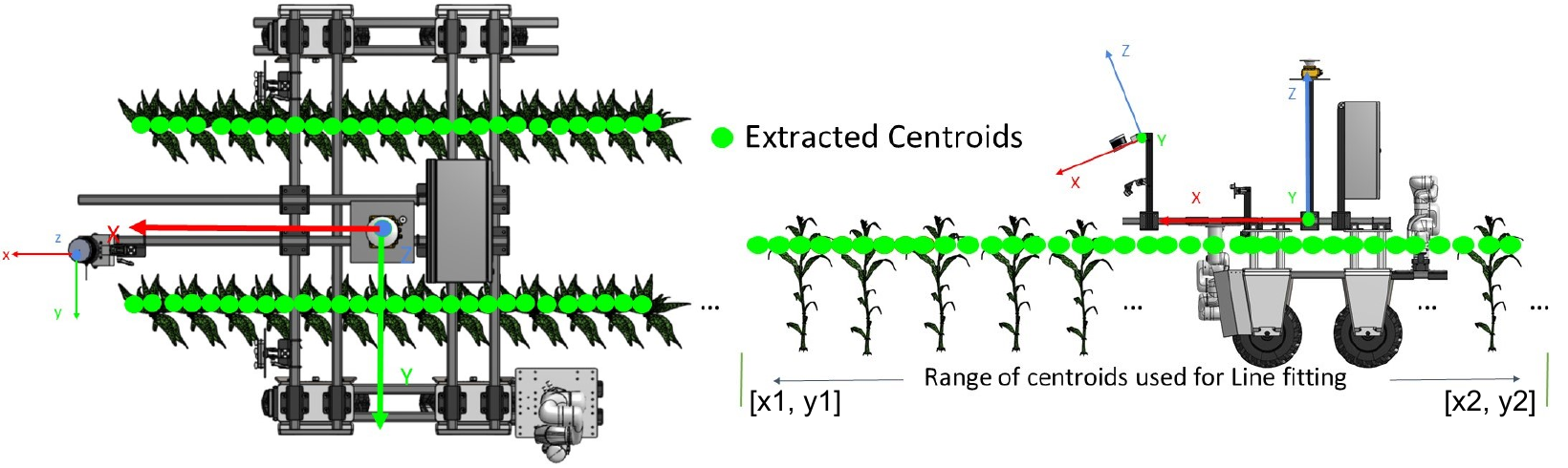} 
  \caption{Illustration of the RANSAC line fitting algorithm for detecting crop rows. Centroids of the first row on the left and right in the robot's frame are extracted (left). The RANSAC line fitting algorithm is then applied (right) between the current farthest detected centroids and previous centroids within a specified range (e.g., 0.5 meters behind the robot). The predicted crop row is in the form of $[x_1, y_1, x_2, y_2]$.}
  \vspace{-1.1em}
  \label{fig:line}
\end{figure}
\begin{figure}[b]
\vspace{-1em}
    \centering
    \includegraphics[width =0.48\textwidth]{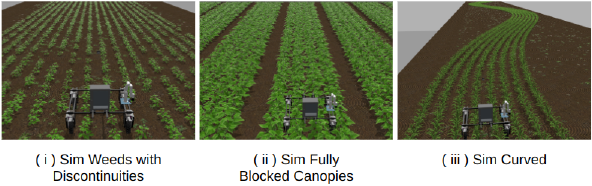}
    \caption{Gazebo simulated fields with three challenging scenarios ((\romannumeral 1) Weeds and Discontinuities, (\romannumeral 2) Fully Blocked Canopies, and (\romannumeral 3) Curved rows.}
    \label{fig:harshrow}
\end{figure}
\begin{figure*}
\vspace{0.15cm}
\begin{center}
\includegraphics[width=0.99\textwidth, height = 5.85cm]{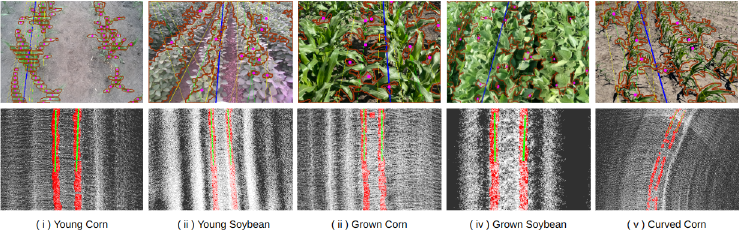}
\end{center}
\caption{Comparison between our method (Bottom, White: Accumulated LiDAR points based on filtered odometry data, Red: Crop centroids, Green: Predicted row) and baseline method (Top, Blue: Navigation line, Green: Predicted row, Yellow: Crop bounding box, Pink: Crop centroids, Brown: Crop contour) in crop row detection. The results show that our method exhibits superior robustness and versatility compared to vision-based systems across different crop types, growth stages, and row trends.}
\vspace{-1.1em}
\label{fig:realdata1}
\end{figure*}

\section{Autonomous Navigation In Row-crop Fields}
The autonomous navigation system within crop rows includes crop-row detection and crop-row following algorithms. Sec. \ref{sec:detection} describes how we extract the raw LiDAR point clouds and perform crop-row detection on the extracted results. Sec. \ref{sec:following} introduces the nonlinear local Model Predictive Control (MPC) algorithm for crop-row following.  

\subsection{Multiple Crop-Row Detection Algorithm}\label{sec:detection}
To leverage the height differences between crops and inter-row spacing, we chose to utilize LiDAR data to perform crop-row detection. LiDAR provides accurate 3D data that better describes crop row patterns when compared with cameras, especially when crop canopy covers the inter-row spacing. The LiDAR used in this work was installed in the center front of the robot and has the same coordinate frame as the robot when facing the horizon. The LiDAR was tilted down along its Y-axis(leftward) to obtain an adequate amount of points for crop-row detection. Based on what we observe in several real fields, we assume that crop rows are planted in parallel and the ground is relatively flat. 

Our crop-row detection algorithm is comprised of three key components. First, we estimate the ground plane using the LiDAR's tilted angle $\theta$, raise it to intersect the point cloud centroid, and filter out points below this plane. This process isolates the returns corresponding only to the top of the plants. Second, employing this filtered LiDAR data, we apply the K-means clustering algorithm to segment crop rows autonomously. The generated centroids of these segments represent the center of crop rows. We utilize the robot's filtered odometry, $[x, y, \psi]$, to accumulate detected centroids into the robot frame within a short time window. This approach allows us to have accurate local positioning and avoid drifting. Finally, we implement the RANSAC line fitting algorithm on this crop-row centroids map, extracting 2D line locations for the first row on the left and the first row on the right in the robot frame. 

\subsubsection{LiDAR Data Preprocessing}
 We employ a multi-step preprocessing approach to identify each crop row within the LiDAR field of view. First, we retain only the points within the frontal $120 \degree$ with a range up to 4m. To exploit the height difference between crops and inter-row spacing for crop-row detection, we generate a virtual plane that intersects the centroid of the point clouds. As shown in Fig. \ref{fig:plane}, this virtual plane, parallel to the ground in the LiDAR frame, is generated based on the known LiDAR tilted angle $\theta$ along the LiDAR Y-axis. We then filter the points located below this plane as: 
\begin{equation}
    {v}_{\perp} = R_{y, \theta} \cdot [0 \ 0 \ 1]^T \quad;\quad 
    {d} = {v}_{\perp} \cdot \overline{P}_{x,y,z}
\end{equation}
\begin{equation}
     {P}_{\text{f}} = \{p \,|\, p_x{v}_{\perp,x} + p_y{v}_{\perp,y} + p_z{v}_{\perp,z} + d > 0\} 
\end{equation}

We determine the normal vector of the ground plane in the LiDAR frame ${v}_{\perp}$ by multiplying the rotation matrix $R_{y, \theta}$ of the ground plane in the LiDAR frame and the unit normal vector $[0, 0, 1]^T$. To elevate the ground plane to intersect the centroid of the LiDAR point clouds, we calculate the elevating distance $d$ along the normal vector of the ground plane by multiplying ${v}_{\perp}$ and the centroid of the point clouds $\overline{P}_{x,y,z}$. We finally filter out all the points below the elevated ground plane and obtain filtered points $ {P}_{\text{f}}$. 

\subsubsection{K-means Clustering and Localization Integration}
We divide LiDAR points into bins for each 1m depth range and apply the K-means algorithm to calculate the centroid ${Q}_{\text{i}}$ for each bin, $i \in \{1, \ldots, n\}$, where $n$ is the number of detected clusters. We store these centroids in each time step into ${Q}_{\text{local}}$ within the current robot frame based on the filtered robot odometry data. We apply a large $n$ initially and merge centroids that are closer than $0.3 m$ so that the final number of centroids $n$ indicates the number of detected crop rows, with each centroid near the row's center. 

\subsubsection{RANSAC Line Fitting}

Due to the sparsity of centroids detected at each time frame, we track centroids' local positions from previous frames to enhance the smoothness and accuracy of currently predicted crop rows. When the distance between detected centroids on each crop row exceeds 2 meters, we apply the RANSAC line fitting (with 1000 iterations) to estimate the position and orientation of each crop row. As the robot progresses, we keep updating the detected centroids for RANSAC line fitting by considering only the centroids between the farthest detected points and those located 0.5 meters behind the robot on each crop row as shown in Fig. \ref{fig:line}. To avoid confusion with the outer rows (the ones further to the left or right), we only extract the centroids' local positions of the first row on the left and the first row on the right in the robot's frame. Finally, the output prediction of each crop row is parametrized in the form of $[x_1, y_1, x_2, y_2]$.

\begin{table*}[t]
\vspace{0.15cm}
 \caption{Our Crop-Row Detection Performance and Comparison of Crop-Row Following Performance: Our Method vs. Baseline \cite{multi_crop_row}.} 

    \label{tab:tab1}
    \centering
  \begin{tabularx}{0.99\textwidth}{
    >{\centering\arraybackslash}m{1.0cm}|
     >{\centering\arraybackslash}m{1.5cm}
     >{\centering\arraybackslash}m{1cm}
     >{\centering\arraybackslash}m{1cm}
     >{\centering\arraybackslash}m{1cm}
     >{\centering\arraybackslash}m{1cm}
     >{\centering\arraybackslash}m{1cm}|
     >{\centering\arraybackslash}m{1cm}
     >{\centering\arraybackslash}m{1cm}
     >{\centering\arraybackslash}m{1cm}
     >{\centering\arraybackslash}m{1cm}
     >{\centering\arraybackslash}m{1cm}
 }
    \hline \addlinespace[0.5ex] 
    \centering & &\multicolumn{5}{c}{\underline{Simulated fields}} & \multicolumn{5}{c}{\underline{Real fields}} \\
    \addlinespace[0.5ex] 
    \centering Crop  & &\centering Young Corn &  \centering Grown Corn &  \centering Young Soybean & \centering Grown Soybean &  Curved Corn &  \centering Young Corn &  \centering Grown Corn &  \centering Young Soybean &\centering Grown Soybean &  Curved Corn \arraybackslash \\

    \hline
    \hline
    \addlinespace[0.5ex] 
    
    \multirow{6}{1.0cm}{Row-Detection}& Length & 200m & 200m & 200m & 200m & 200m & 88m & 85m & 207m & 200m & 200m \\
    \addlinespace
     &\(\mu \pm \sigma\) of dist error & 1.67 \(\pm \) 1.12cm & 3.74 \(\pm \) 2.49cm & 1.54 \(\pm \) 1.07cm & 3.04 \(\pm \) 1.69cm & 2.72 \(\pm \) 1.33cm & 2.97 \(\pm \) 2.45cm & 3.75 \(\pm \) 2.98cm & 4.83 \(\pm \) 3.25cm & 5.73 \(\pm \) 3.82cm & 3.46 \(\pm \) 2.33cm \\
    \addlinespace[0.5ex] 
    &\(\mu \pm \sigma\) of angular error & 0.28 \(\pm \) 0.23\degree & 1.77 \(\pm \) 1.51\degree & 0.49\(\pm \) 0.33 \degree & 1.14\(\pm \) 1.01 \degree & 0.92 \(\pm \) 0.71 \degree & 1.42 \(\pm \) 1.23\degree & 2.26 \(\pm \) 1.89\degree & 2.71 \(\pm \) 1.85\degree & 3.85 \(\pm \) 2.43\degree & 2.73 \(\pm \) 1.91\degree \\
    \addlinespace[0.5ex]
    \hline
    \hline
    \addlinespace[0.5ex]
    \centering Crop   & &\centering Young Corn &  \centering Grown Corn &  \centering Young Soybean & \centering Grown Soybean &  Curved Corn &  \multicolumn{3}{c}{\centering Young Corn} &  \multicolumn{2}{c}{\centering Grown Corn} \arraybackslash \\
    \addlinespace[0.5ex]
    \hline
    \hline
    \addlinespace[0.5ex]
    \multirow{5}{1.0cm}{Row-Following}& Length & 200m/15m$^{*}$& 200m & 200m/14m$^{*}$ & 200m & 200m & \multicolumn{3}{c}{48m} & \multicolumn{2}{c}{47m}\\
    \addlinespace
     &\multicolumn{1}{c}{\multirow{3}{1.7cm}{\centering \(\mu \pm \sigma\) of dist to center line}} & 0.97 \(\pm \) 0.79cm & 2.45 \(\pm \) 1.69cm & 0.64 \(\pm \) 0.37cm & 1.24 \(\pm \) 0.97cm & 9.21 \(\pm \) 7.25cm & \multicolumn{3}{c}{4.76 \(\pm \) 3.25cm} & \multicolumn{2}{c}{5.58 \(\pm \) 4.06cm} \\
    \addlinespace[0.5ex] 
    Baseline& & 22.4 \(\pm \) 7.79cm & N/A & 23.5 \(\pm \) 8.01cm & N/A & N/A & \multicolumn{3}{c}{N/A} & \multicolumn{2}{c}{N/A} \\
    \addlinespace[0.5ex]
    \hline
    
\end{tabularx}
\begin{tablenotes}
            \small
            \item[**] $^{*}$ The baseline crop-row following method fails at 15m and 14m in simulated young corn and young soybean.
            \item[**] \hspace{0.28cm}N/A: The baseline method fails in detecting crop rows.
        \end{tablenotes}
\vspace{-1.2em}
\end{table*}
\begin{figure}[t]
    \centering
    \includegraphics[width =0.48\textwidth, height = 4.8cm]{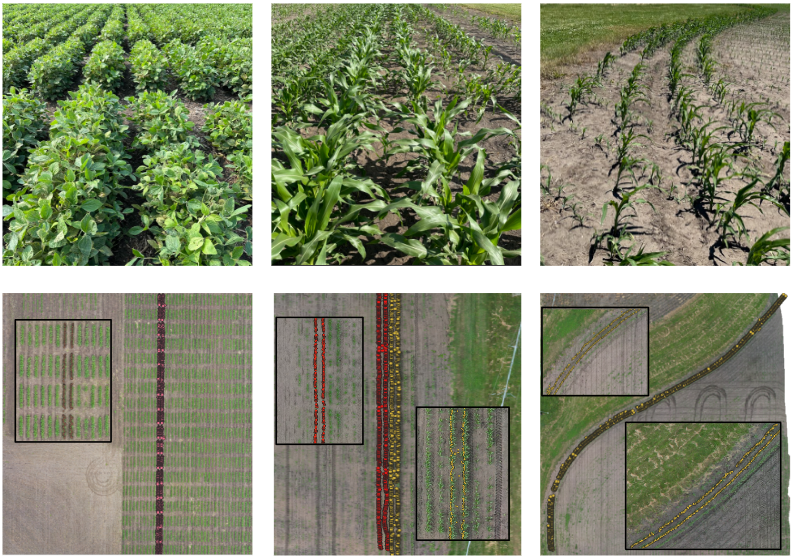}
    \caption{Fields overview and the collected drone maps. The experiments are conducted in soybean (top left), corn (top middle), and curved corn (top right) fields. We collected the drone map for (soybean (bottom left), corn (bottom middle), and curved corn (bottom right)) with RTK-GPS inputs as ground truth for later evaluations. We overlay the detected centroids on the drone map to provide qualitative results, demonstrating that the detected centroids align with the crop rows as shown in the map.}
    \vspace{-1em}
    \label{fig:drone}
\end{figure}

\subsection{Multiple Crop-row Following System}\label{sec:following}
After crop-row detection, we generate waypoints along the center line of the two predicted crop rows. We apply a nonlinear Model Predictive Control (MPC) algorithm in the robot's local frame for tracking the generated waypoints. ACADO \cite{acado} is used to solve the quadratic programming problem, enabling real-time operation. 

\section{Experimental Results and Discussion}
To validate the capability and robustness of our crop-row detection, crop-row following, and crop-row switching algorithm, we perform three experiments separately in both Gazebo simulated fields and real fields. The simulation environments are built in the Gazebo simulator with different crop types (corn and soybean) and growth stages(young and grown) by using the Cropcraft tool \cite{cropcraft}. As shown in Fig. \ref{fig:harshrow}, we also implement three challenging scenarios—weeds with discontinuities, fully covered canopies, and curved crop rows—to evaluate our navigation system's performance under realistic agricultural conditions. The real-world experiments were conducted in corn and soybean fields at various stages of crop growth as shown in Fig. \ref{fig:drone}. For both simulated fields and real fields, the inter-row spacing is $0.75m$. As noted in Sec. \ref{sec:robot_platform}, the robot's width is $1.8m$, allowing a tolerance of $22.5cm$ when navigating across two crop rows without causing damage to crops.

In simulated fields, we implement an Amiga robot model on a 1:1 scale to better simulate and evaluate the navigation performance. In both simulation and reality, the LiDAR is positioned at a height of $1.5m$ above the ground and a tilted angle of $30\degree$. We utilize the cuML package (GPU-accelerated) to ensure real-time operation on the K-means clustering algorithm while maintaining compatibility with CPU-only machines. 


\subsection{Multi-Crop-Row Detection}
In the first experiment, we evaluate the robustness of the crop-row detection algorithm under diverse field conditions. These experiments were conducted using the Amiga robot in both Gazebo-simulated and real-world fields, including corn and soybeans at different growth stages. The simulated fields were constructed with total dimensions of $200m$ × $12m$, comprising 16 rows with weeds and discontinuities.
 \begin{figure}[t]
  \centering
  \includegraphics[width=0.48\textwidth]{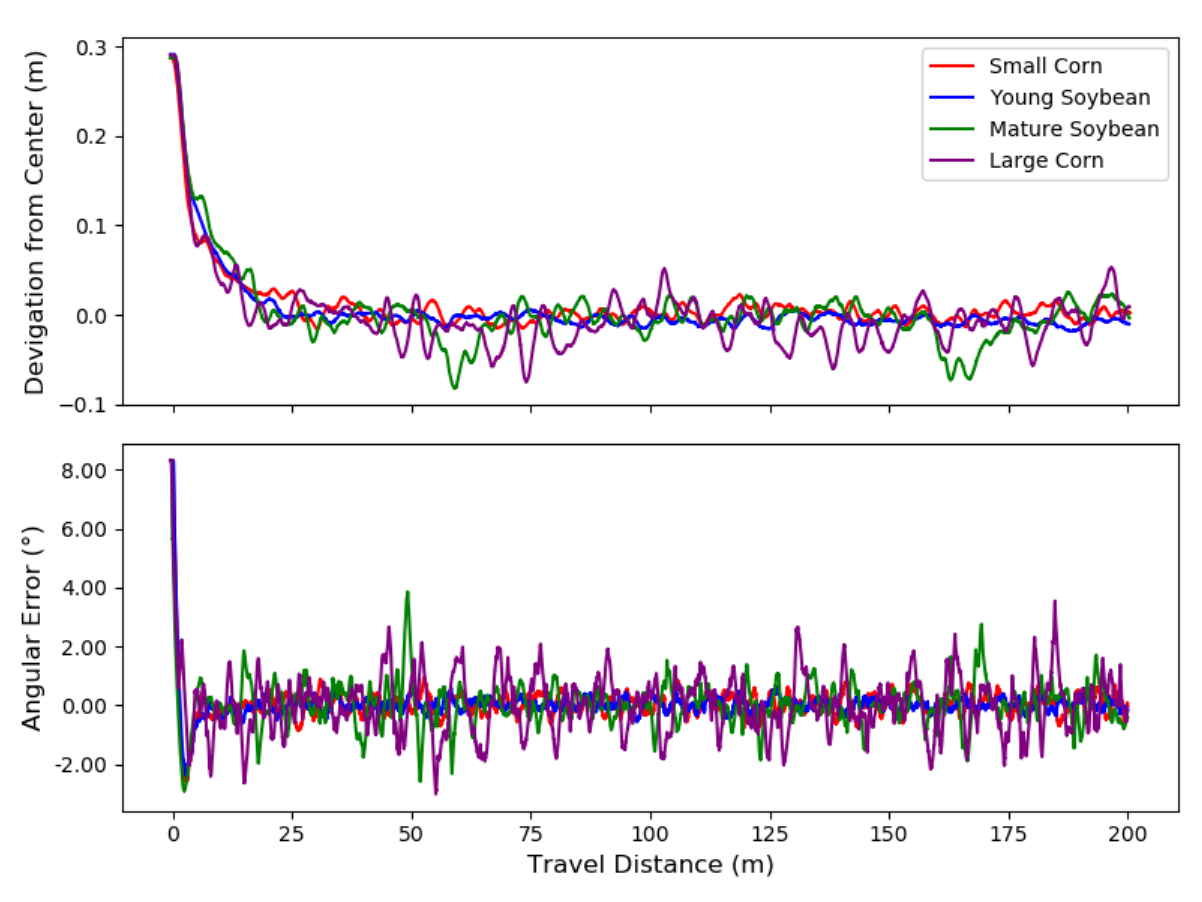} 
  \caption{The distance and angular error plot during autonomous navigation in simulated fields. Distance error (top) and Angular error (bottom) are plotted along the robot move forward distances. The robot successfully recovers from an initial distance error of $0.3m$ and an angular error of $8\degree$.}
  \vspace{-1em}
  \label{fig:results}
\end{figure}
During the experiments, a human operator controls the Amiga robot through the agricultural field while the crop-row detection algorithm operates continuously. 
To assess the accuracy of the detected centroids, ground truth positions of both the crops and the Amiga robot are necessary. In the simulated field, these positions are easily derived. In real fields, drone-generated maps are used (shown in Fig. \ref{fig:drone}) to obtain the ground truth crop positions, while the robot's ground truth position is determined from filtered odometry data $[x,y,\psi]$ \cite{robot-localization} supplemented by RTK-GPS inputs. The performance of the crop-row detection algorithm is evaluated using mean absolute error (MAE) with standard deviation for distance and angular errors between predicted rows and ground truth.

Based on results in Table \ref{tab:tab1}, our method achieves an average detection accuracy of $3.35cm$ when predicting the crop row positions across various simulated and actual agricultural fields. Similarly, the prediction of crop row orientation achieves an average accuracy of $1.76\degree$ across the same fields. The grown corn and soybean fields pose the most challenging scenario for crop-row detection due to the blocking of inter-row spacing by their canopies. The baseline crop-row detection algorithms fail for this reason \cite{multi_crop_row}. Our method successfully detects the grown soybean with an average accuracy of $5.73cm$ and the grown corn with an average accuracy of $3.75cm$ in real fields. Fig \ref{fig:realdata1} shows the comparisons between our method and baseline method \cite{multi_crop_row} across different crop types, growth stages, and row trends. While the baseline method needs parameter fine-tuning for each crop and only works on young crops, our method outperforms it across all testing fields without requiring parameter adjustments. These qualitative results showcase the feasibility of our crop-row detection algorithm in detecting various crops in the actual fields.  A potential limitation of our crop-row detection approach arises when crops are in the germination stage, leading to significant gaps between crops and minor height differences between crops and the ground. Despite this, our crop-row detection approach demonstrates its ability to detect multiple crop rows in various challenging agricultural fields reliably.

\subsection{Multi-Crop-Row Following}
\begin{figure}[t]
\vspace{0.15cm}
  \centering
  \includegraphics[width=0.48\textwidth, height=4cm]{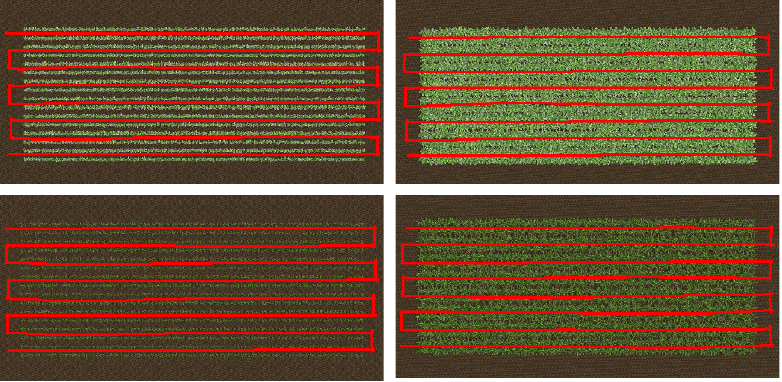} \\
  \caption{Trajectory (red) of the robot while navigating across the whole field (green) in Gazebo-simulated young soybean (top left), young corn (bottom left), grown soybean (top right), and grown corn (bottom right) fields ($30m$ × $12m$ with $16$ rows).}
  \vspace{-1.2em}
  \label{fig:switching}
\end{figure}
To evaluate the performance of the crop-row following algorithm, we perform experiments in the same fields. During these tests, both crop-row detection and following algorithms are activated. Once the detection algorithm predicts the crop row positions and orientations, the crop-row following algorithm will control the Amiga robot to follow the detected crop rows without human intervention.

The optimal performance during crop row navigation is achieved by maintaining the robot at the center line between the crop rows. The ground truth positions of the robot and the center line between crop rows are derived using the same method as in the crop-row detection experiments. We apply the same mean absolute error (MAE) with standard deviation as the metric to evaluate the crop-row following performance, quantifying the error between robot odometry and the desired center line. The results are presented in Table \ref{tab:tab1}.

A suboptimal crop-row following algorithm, causing oscillations during navigation, can lead to failures in the crop-row detection algorithm to track the crop rows accurately and thus make the whole system fail, such as visual servoing (shown in Table \ref{tab:tab1}). As detailed in Table \ref{tab:tab1}, our navigation system successfully performs autonomous navigation within the crop rows, achieving an average deviation of $3.55cm$ ($5\%$ inter-row spacing) from the center line between the crop rows in all fields without human intervention and without harming the crops. Additionally, as shown in Fig. \ref{fig:results}, the system effectively recovers from an initial distance error of $0.3m$ or an angular error of $8\degree$. These results, demonstrating minimal deviation from the ideal crop row path and effective recovery, underscore the viability and robustness of our navigation system as a crop-agnostic approach for autonomous navigation in diverse real-world agricultural environments.
\subsection{Multi-Crop-Row Switching Performance}
The final experiment assesses the performance of our navigation system in managing crop-row transitions throughout an entire field. We perform experiments in the Gazebo simulated fields. We generate the experiment fields with a dimension of $30m$ × $12m$ with an inter-row spacing of $0.75m$. The robot is tasked with traversing from the initial two rows in the top-left corner to the final two rows in the bottom-right corner. As shown in Fig. \ref{fig:switching}, the robot successfully covers the whole field with our crop-row switching strategy. At the end of each row, the robot performs the crop-row switching maneuver within a $1.5m$ space, demonstrating our navigation system's capability and robustness in navigating complex fields with limited maneuvering space. 
\section{Conclusion}
In this paper, we present a novel LiDAR-based crop-row detection approach that integrates the Model Predictive Control (MPC) and lane-switching algorithm to create an autonomous navigation system for agricultural robots in row-crop fields. This system facilitates independent robot navigation for diverse agricultural tasks, contributing to precision farming. Our crop-row detection method utilizes 3D LiDAR data to extract the height information and accurately detects crop rows amidst challenging scenarios such as canopy obstructions. The whole navigation system incorporates the crop-row detection, following, and switching algorithm, enabling automated tracking of detected crop rows and full field coverage. This navigation system is evaluated in both actual fields and Gazebo simulated fields with a 1:1 scale Amiga robot model. The crop-row detection algorithm achieves an average detection accuracy of $3.35cm$, while the crop-row following algorithm achieves an average driving accuracy of $3.55cm$. Future work will focus on improving the robustness of the crop row perception algorithm by integrating camera data, especially to handle gaps between plants during the germination stage.
\section{Acknowledgments}
We would like to thank Iowa State University for providing field research opportunities at the Curtiss farm and Agronomy farm. This work was supported by: NSF/USDA-NIFA AIIRA AI Research Institute 2021-67021-35329.

\bibliographystyle{IEEEtran}
\bibliography{IEEE}

\end{document}